\documentclass{article}

\usepackage{arxiv}

\usepackage[utf8]{inputenc} % allow utf-8 input
\usepackage[T1]{fontenc}    % use 8-bit T1 fonts
\usepackage{hyperref}       % hyperlinks
\usepackage{url}            % simple URL typesetting
\usepackage{booktabs}       % professional-quality tables
\usepackage{amsfonts}       % blackboard math symbols
\usepackage{nicefrac}       % compact symbols for 1/2, etc.
\usepackage{microtype}      % microtypography
\usepackage{lipsum}
\usepackage{graphicx}
\usepackage{comment}
\usepackage{multirow}
\usepackage{subcaption}
\usepackage{amsmath}
\usepackage[psamsfonts]{amssymb}
\usepackage{amsxtra}
\usepackage{threeparttable}
\title{E-PixelHop: An Enhanced PixelHop Method \\for Object Classification}

\author{
  Yijing Yang \\
  University of Southern California\\
  Los Angeles, California, USA \\
  \texttt{yijingya@usc.edu} \\
     \And
  Vasileios Magoulianitis \\
  University of Southern California\\
  Los Angeles, California, USA \\
  \texttt{magoulia@usc.edu} \\
   \And
  C.-C. Jay Kuo \\
  University of Southern California\\
  Los Angeles, California, USA \\
  \texttt{jckuo@usc.edu} \\

}

\begin{document}
\maketitle

\begin{abstract}

Based on PixelHop and PixelHop++, which are recently developed using the
successive subspace learning (SSL) framework, we propose an enhanced
solution for object classification, called E-PixelHop, in this work.
E-PixelHop consists of the following steps. First, to decouple the color
channels for a color image, we apply principle component analysis and
project RGB three color channels onto two principle subspaces which are
processed separately for classification. Second, to address the
importance of multi-scale features, we conduct pixel-level
classification at each hop with various receptive fields. Third, to further improve pixel-level classification
accuracy, we develop a supervised label smoothing (SLS) scheme to ensure
prediction consistency.  Forth, pixel-level decisions from each hop and
from each color subspace are fused together for image-level decision.
Fifth, to resolve confusing classes for further performance boosting, we
formulate E-PixelHop as a two-stage pipeline.  In the first stage,
multi-class classification is performed to get a soft decision for each
class, where the top 2 classes with the highest probabilities are called
confusing classes. Then, we conduct a binary classification in the
second stage. The main contributions lie in Steps 1, 3 and 5.  We use
the classification of the CIFAR-10 dataset as an example to demonstrate
the effectiveness of the above-mentioned key components of E-PixelHop. 

\end{abstract}

\section{Introduction}\label{sec:introduction}

Object classification has been studied for many years as a fundamental
problem in computer vision. With the development of convolutional neural
networks (CNNs) and the availability of larger scale datasets, we see a
rapid success in the classification using deep learning for both low-
and high-resolution images \cite{lecun1998gradient}, \cite{he2016deep},
\cite{huang2017densely}, \cite{simonyan2014very},
\cite{szegedy2015going}. Deep learning networks use backpropagation to
optimize an objective function to find the optimal parameters of
networks. Although being effective, deep learning demands a high
computational cost. As the network goes deeper, the model size increases
dramatically.  Research on light-weight neural networks
\cite{iandola2016squeezenet}, \cite{howard2017mobilenets},
\cite{tan2019efficientnet} has received attention to address the
complexity issue. One major challenge associated with deep learning is
that its underlying mechanism is not transparent. 

Recently, based on the successive subspace learning (SSL) framework, the
PixelHop \cite{chen2020pixelhop} and the PixelHop++
\cite{chen2020pixelhop++} methods have been proposed for image
classification. Both follow the traditional pattern recognition paradigm
and partition the classification problem into two cascaded modules: 1)
feature extraction and 2) classification. Powerful spatial-spectral
features can be extracted from PixelHop and PixelHop++ in an
unsupervised manner. Then, they are fed into a trained classifier
for final decision. Every step in PixelHop/PixelHop++ is explainable,
and the whole solution is mathematically transparent.

In this work, we propose an enhanced PixelHop method, named E-PixelHop.
It consists of the following main steps. 
\begin{enumerate}
\item To decouple the color channels for a color image, E-PixelHop
applies principle component analysis (PCA) and project RGB three color
channels onto two principle subspaces which are processed separately for
classification. 
\item To address the importance of multi-scale features, we conduct
pixel-level classification at each hop which corresponds to a patch of
various sizes in the input image. 
\item To further improve pixel-level classification accuracy, we develop
a supervised label smoothing (SLS) scheme to ensure intra-hop and
inter-hop pixel-level prediction consistency. 
\item Pixel-level decisions from each hop and from each color subspace
are fused together for image-level decision. 
\item To resolve confusing classes for further performance boosting, we
formulate E-PixelHop as a two-stage pipeline.  In the first stage,
multi-class classification is performed to get a soft decision for each
class, where the top 2 classes with the highest probabilities are called
confusing classes. 
\end{enumerate}
The main contributions of E-PixelHop lie in Steps 1, 3 and 5.  

The rest of this paper is organized as follows. Related work is reviewed
in Sec. \ref{sec:review}. The E-PixelHop method is presented in Sec.
\ref{sec:method}. The label smoothing procedure is detailed in Sec.
\ref{sec:smoothing}. Experimental setup and results are detailed in
Sec. \ref{sec:experiments}. Finally, concluding remarks and future
research directions are drawn in Sec. \ref{sec:conclude}. 

\section{Review of Related Work}\label{sec:review}

\subsection{Multi-scale Features for Object Classification} 

Handcrafted features were extracted before the deep learning era.  To
obtain multi-scale features, Mutch {\em et al.}
\cite{mutch2006multiclass} applied Gabor filters to all positions and
scales. Scale-invariant features can be derived by alternating template
matching and max-pooling operations.  Schnitzspan {\em et al.}
\cite{schnitzspan2008hierarchical} proposed a hierarchical random field
that combines the global-feature-based methods with the
local-feature-based approaches in one consistent multi-layer framework.
In deep learning, the decision is usually made using features from the
deepest convolutional layer. Recently, more investigations are made to
exploit outputs from shallower layers to improve the classification
performance.  For example, Liu {\em et al.} \cite{liu2015treasure}
proposed a cross-convolutional-layer pooling operation that extracts
local features from one convolutional layer and
pools the extracted features with the guidance of the next convolutional
layer. Jetley {\em et al.} \cite{jetley2018learn} extracted features
from shallower layers and combined them with global features to estimate
attention maps for further classification performance improvement, where
global features are used to derive attention in local features that are
consistent with the semantic meaning of the underlying objects. 

\subsection{Hierarchical Classification Strategy} 

It is easier to distinguish between classes of dissimilarity than those
of similarity. For example, one should distinguish between cats and cars
better than between cats and dogs. Along this line, one can build a
hierarchical relation among multiple classes based on their semantic
meaning to improve classification performance \cite{liu2017easy,
seo2019hierarchical, zweig2007exploiting}. Alsallakh {\em et al.}
\cite{bilal2017convolutional} investigated ways to exploit the
hierarchical structure to improve classification accuracy of CNNs.  Yan
{\em et al.} \cite{yan2015hd} proposed a hierarchical deep CNN (HD-CNNs)
which embeds deep CNNs into a category hierarchy. It separates easy
classes using a coarse category classifier and distinguishes difficult
classes with a fine category classifier. Chen {\em et al.}
\cite{chen2017exploring} proposed to merge images associated with
confusing anchor vectors into a confusion set and split the set to create multiple subsets in an
unsupervised manner. A random forest
classifier is then trained for each confusion subset to boost the
scene classification performance. An identification and resolution scheme of
confusing categories was proposed by Li {\em et al.}
\cite{li2018improving} based on binary-tree-structured clustering.  It
can be applied to any CNN-based object classification baseline to
further improve its performance. Zhu {\em et al.} \cite{zhu2017b} defined different levels of categories and proposed a Branch
Convolutional Neural Network (B-CNN) that outputs multiple predictions
and ordered them from coarse to fine along concatenated convolutional
layers. This procedure is analogous to a hierarchical object
classification scheme and enforces the network to learn human understandable concepts in different layers.

\subsection{Successive Subspace Learning} 

Being inspired by deep learning, the successive subspace learning (SSL)
methodology was proposed by Kuo {\em et al.} in a sequence of papers
\cite{kuo2016understanding, kuo2017cnn, kuo2018data,
kuo2019interpretable}. SSL-based methods learn feature representations
in an unsupervised feedforward manner using multi-stage principle
component analysis (PCA).  Joint spatial-spectral representations are
obtained at different scales through multi-stage transforms.  Three
variants of the PCA transform were developed. They are the Saak
transform \cite{kuo2018data, chen2018saak}, the Saab transform
\cite{kuo2019interpretable}, and the channel-wise (c/w) Saab transform
\cite{chen2020pixelhop++}.  The c/w Saab transform is the most effective
one among the three since it can reduce the model size and improve the
transform efficiency by learning filters in each channel separately.
This is achieved by exploiting the weak correlation between different
spectral components in the Saab transform.  These features can be used
to train classifers in the training phase and provide inference in the
test phase. Two object classification pipelines, PixelHop
\cite{chen2020pixelhop} using the Saab transform and PixelHop++
\cite{chen2020pixelhop++} using the c/w Saab transform, were designed.
Since the feature extraction module is unsupervised, a label-assisted
regression (LAG) unit was proposed in PixelHop as a feature
transformation, aiming at projecting the unsupervised feature to a more
separable subspace.  In PixelHop++, cross-entropy based feature
selection is applied before each LAG unit to select task-dependent
features, which provides a flexible tradeoff between model size and
accuracy.  Ensemble methods were used in \cite{chen2019ensembles} to
boost the classification performance.  SSL has been successfully applied
to many application domains. Examples include~\cite{zhang2020pointhop++,
zhang2020pointhop, zhang2020unsupervised, kadam2020unsupervised,
manimaran2020visualization, tseng2020interpretable,
rouhsedaghat2020facehop, lei2020nites, rouhsedaghat2021successive,
chen2021defakehop, zhang2021anomalyhop, kadam2021r, liu2021voxelhop}. 

\section{E-PixelHop Method}\label{sec:method}
%%%%%%%%%%%%%%%%%%%%%%%%%%%%%%%%%%%%%%%%%%%%%%%%%%%%%%%%%%
\begin{figure*}[t]
\begin{center}
\includegraphics[width=0.85\linewidth]{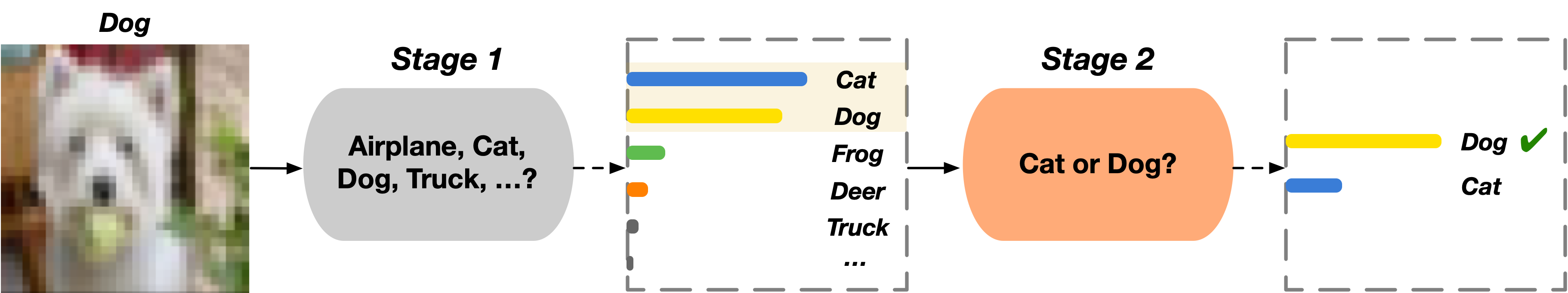}
\end{center}
\caption{E-PixelHop is a two-stage sequential classification method.}
\label{fig:twostage}
\end{figure*}
%%%%%%%%%%%%%%%%%%%%%%%%%%%%%%%%%%%%%%%%%%%%%%%%%%%%%%%%%%

The E-PixelHop method for object classification is proposed in this
section. An overview of E-PixelHop is described in Sec.
\ref{subsec:overview}.  Then, a multi-class classification baseline is
described in Sec.  \ref{subsec:baseline}.  Finally, confusion class
resolution is presented in Sec.  \ref{subsec:confuse}. 

\subsection{System Overview}\label{subsec:overview}

The overall framework of the E-PixelHop is illustrated in Fig.
\ref{fig:twostage}. It is a two-stage sequential classification method.
The first stage serves as a baseline that performs classification among
all object classes. Its output is a set of soft decision scores that
indicate the probability of each class. The classes with the highest $M$
probabilities for each image as its confusing group. Typically, we set
$M=2$. In the second stage, a one-versus-one competition is conducted to
refine the prediction result.  The two-stage sequential decision makes a
coarse prediction and, then, focuses on confusing classes resolution for
more accurate prediction. 

We follow the traditional pattern recognition paradigm by dividing the
problem into feature extraction and classification two separate modules.
Multi-hop c/w Saab transforms are used to extract joint spatial-spectral
features in an unsupervised manner.  For classification, we adopt
pixel-based classification in each hop that predicts objects of
different scales at different spatial locations, where a pixel in a
deeper hop denotes a patch of a larger size in the input image.  In the
training, ground truth labels at pixels follow image labels.
Pixel-based classification based on the intra-hop information only tends
to be noisy. To address this problem, a supervised label smoothing (SLS)
method is proposed to reduce prediction uncertainty. 

%%%%%%%%%%%%%%%%%%%%%%%%%%%%%%%%%%%%%%%%%%%%%%%%%%%%%%%%%%
\begin{figure*}[h]
\begin{center}
\includegraphics[width=0.97\linewidth]{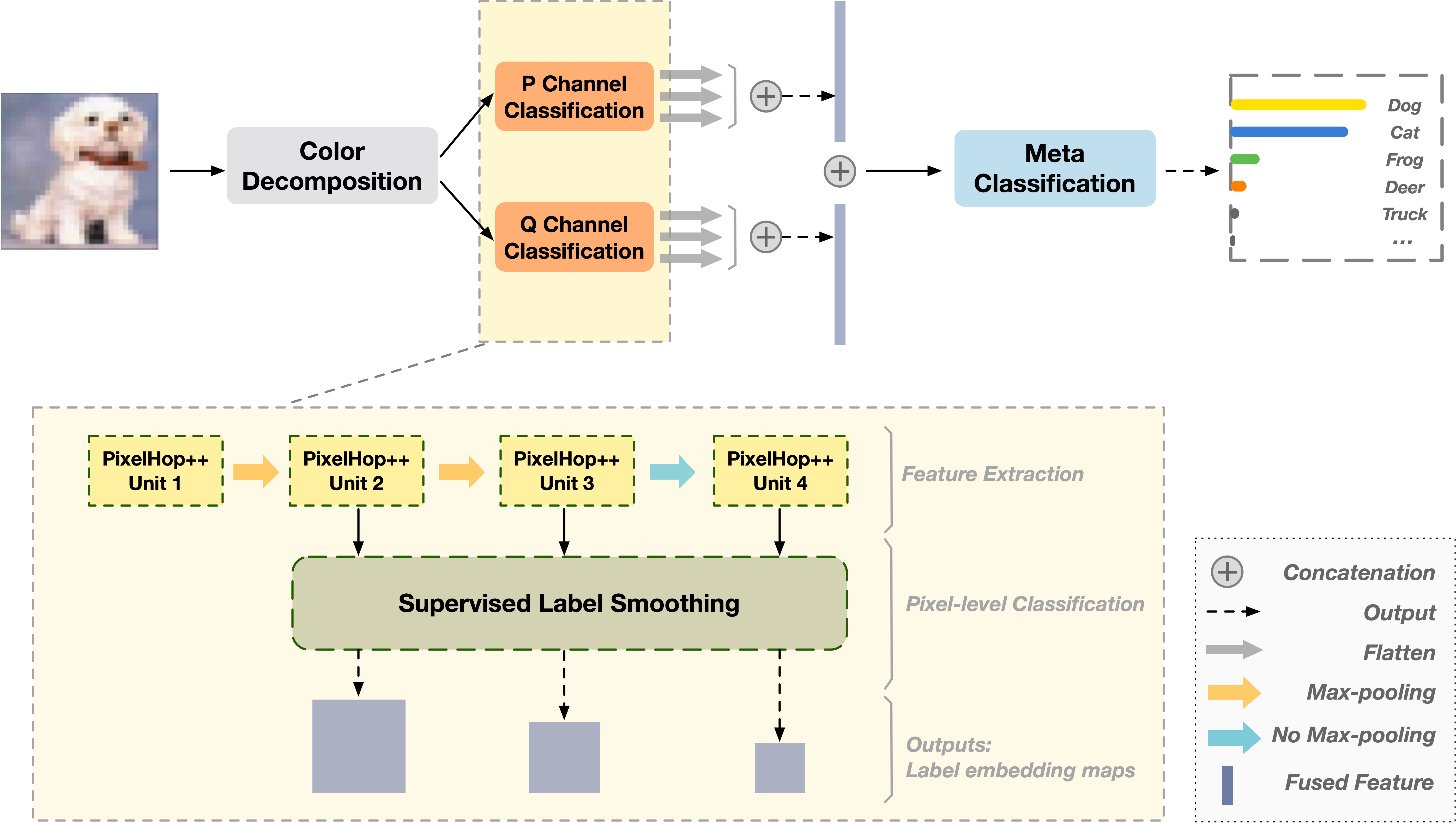}
\end{center}
\caption{Illustration of the E-PixelHop classification baseline.}
\label{fig:baseline}
\end{figure*}
%%%%%%%%%%%%%%%%%%%%%%%%%%%%%%%%%%%%%%%%%%%%%%%%%%%%%%%%%%

\subsection{Stage 1: Multi-class Baseline Classifier}\label{subsec:baseline}

We use the CIFAR-10 dataset \cite{cifar10} as an example to explain the
architecture of the multi-class classification baseline. It consists of
four modules: 1) Color decomposition, 2) PixelHop++ feature extraction,
3) Pixel-level classification with label smoothing, 4) Meta
classification. The system diagram for the baseline classification is
shown in Fig. \ref{fig:baseline}. 

\textbf{1) Color decomposition.} For input color images of RGB three
channels, the feature dimension is three times of the spatial dimension
in a local neighborhood. To reduce the dimension, we apply the principle
component analysis (PCA) to the 3D color channels of all pixels. That
is, we collect RGB 3D color vectors from pixels in all input images and
learn the PCA kernels from collected samples. They project the RGB color
coordinates to decoupled color coordinates, which are named the
\textit{PQR} color coordinates, where P and Q channels correspond to the
$1^{st}$ and $2^{nd}$ principle components. Experimental results show
that P and Q channels contain approximately 98.5\% of the energy of RGB
three channels. Specifically, the P channel has 91.08\% and the Q
channel 7.38\%. Since P and Q channels are uncorrelated with each other,
we can proceed them separately in modules 2 and 3.

\textbf{2) PixelHop++ feature extraction.} As shown in Fig.
\ref{fig:baseline}, we use multiple PixelHop++ units
\cite{chen2020pixelhop++} to extract features from P and Q channels,
respectively. One PixelHop++ unit consists of two cascaded operations: 1)
neighborhood construction with a defined window size in the spatial
domain, and 2) the channel-wise (c/w) Saab transform.  The input to the
first PixelHop++ unit is a tensor of dimension $(S_0\times S_0) \times
K_0$, where $S_0\times S_0$ is the spatial dimension, $K_0$ is the
spectral dimension, and subscript $0$ indicates that it is the hop-0
representation. For raw RGB images, $K_0=3$.  Through color transformation, we have
$K_0=1$ for each individual P or Q channel. The output of the $i$-th
PixelHop++ unit is denoted by $(S_i \times S_i ) \times K_i$, which is
the hop-$i$ representation.  The hyper parameter $K_i$ is determined by
the number of total c/w Saab coefficients kept in hop-$i$.  As $i$
increases, the receptive field associated with a pixel in hop-$i$
becomes larger and more global information of the input image is
captured.  To enable faster expansion of the receptive field, we apply
max-pooling with stride $2$ in shallower hops. 

\textbf{3) Pixel-level classification with label smoothing.} The output
from each PixelHop++ unit can be viewed as a feature tensor.  The
feature vector associated with a pixel location is fed into a
pixel-level classifier to get a probability vector whose element
indicates the probability of belonging to a certain object class.  This
probability vector can be viewed as a soft decision or a soft label.
The soft label can be noisy, and we propose a supervised label smoothing mechanism
to make the soft decision at neighboring spatial locations and hops more
consistent. This is one of the main contributions of E-PixelHop. It
will be elaborated in Sec. \ref{sec:smoothing}.

\textbf{4) Meta classification.} The soft decision maps from different
hops of P and Q channels are concatenated and used to train a meta
classifier, which gives the baseline prediction. 

\subsection{Stage 2: Confusion Set Resolution}\label{subsec:confuse}

Instead of constructing a hierarchical learning structure manually as
done in \cite{zhu2017b}, E-PixelHop identifies confusion sets using the
predictions of the baseline classifier in Stage 1. The baseline
classifier outputs a $C$-dimensional soft decision vector for each
image.  Generally, the $M$ classes with the highest probabilities can
define a confusion set. For a $C$-class classification problem, we have
at most $N_{cg} = \frac{C!}{M!(C-M)!}$ confusion sets.  In practice,
only a portion of $N_{cg}$ sets contribute to the performance gain
significantly since the member of some confusion set is few. In the
following, we focus on the case with $M=2$.  The confusion matrix of the
baseline classifier for CIFAR-10 is shown in Table \ref{tab:cm}.  We see
from the table that ``Cat'' and ``Dog'' are more likely to be confused
with each other. 

Without loss of generality, we use the ``Cat and Dog'' two confusing
classes from CIFAR-10 as an example to explain our approach for
confusion resolution.  In the training, we collect all 5,000 Cat images
and 5,000 Dog images from the training set.  For test images that have
the top-2 candidates of Cat or Dog, we include them in the Cat/Dog
confusion set. If an image belongs to this set yet its ground truth
class is neither Cat nor Dog, its misclassification cannot be corrected.
As a result, the top-2 accuracy of the baseline classifier offers an
upper bound for the final performance.  For binary classification, the
PixelHop++ features learned from the 10 classes in the baseline for P
and Q channels are reused.  Since the label is a 2-D vector whose
element sum is equal to unity, we can simplify the label to a scalar in
label smoothing. Thus, the lab map has a size of $S_i\times S_i \times
1$ at hop-$i$. We use the same ensemble scheme to fuse P and Q predicted
labels. The meta classifier finally yields a binary decision for each
test image in the confusion set to be a dog or a cat.
%%%%%%%%%%%%%%%%%%%%%%%%%%%%%%%%%%%%%%%%%%%%%%%%%%%%%%%%%%
\begin{table*}[t]
\centering
\caption{The confusion matrix for the CIFAR-10 dataset, where the first
row shows the predicted object labels and the first column shows the
ground truth} \label{tab:cm}
\begin{tabular*}{\textwidth}{@{\extracolsep{\fill}}l|cccccccccc}
\hline
           & airplane       & automobile     & bird           & cat            & deer           & dog            & frog           & horse          & ship           & truck          \\ \hline
airplane   & \textbf{0.715} & 0.039          & 0.071          & 0.020          & 0.018          & 0.009          & 0.019          & 0.015          & 0.060          & 0.034          \\
automobile & 0.019          & \textbf{0.852} & 0.008          & 0.005          & 0.006          & 0.004          & 0.010          & 0.007          & 0.015          & 0.074          \\
bird       & 0.059          & 0.006          & \textbf{0.639} & 0.069          & 0.077          & 0.047          & 0.065          & 0.024          & 0.007          & 0.007          \\
cat        & 0.025          & 0.012          & 0.072          & \textbf{0.547} & 0.053          & 0.153          & 0.070          & 0.029          & 0.016          & 0.023          \\
deer       & 0.016          & 0.007          & 0.049          & 0.029          & \textbf{0.694} & 0.045          & 0.060          & 0.089          & 0.007          & 0.004          \\
dog        & 0.012          & 0.012          & 0.040          & 0.199          & 0.042          & \textbf{0.618} & 0.032          & 0.054          & 0.005          & 0.008          \\
frog       & 0.007          & 0.005          & 0.044          & 0.056          & 0.030          & 0.027          & \textbf{0.815} & 0.002          & 0.011          & 0.003          \\
horse      & 0.018          & 0.004          & 0.024          & 0.048          & 0.058          & 0.053          & 0.013          & \textbf{0.768} & 0.005          & 0.009          \\
ship       & 0.047          & 0.025          & 0.014          & 0.008          & 0.004          & 0.005          & 0.008          & 0.005          & \textbf{0.858} & 0.026          \\
truck      & 0.024          & 0.063          & 0.006          & 0.009          & 0.001          & 0.003          & 0.002          & 0.013          & 0.013          & \textbf{0.866} \\ \hline
\end{tabular*}
\end{table*}
%%%%%%%%%%%%%%%%%%%%%%%%%%%%%%%%%%%%%%%%%%%%%%%%%%%%%%%%%%

\section{Supervised Label Smoothing}\label{sec:smoothing}

Soft label prediction at each pixel of a certain hop tends to be noisy.
Label smoothing attempts to convert noisy decision to a clean one using
adjacent label prediction from the same hop or from adjacent hops.  The
uncertainty of intra-hop prediction is first discussed in Sec.
\ref{subsec:intrahop}. The supervised label smoothing (SLS) scheme is
then presented in Sec.  \ref{subsec:SLS}.  

\subsection{Intra-Hop Prediction} \label{subsec:intrahop}

For image classification, only the global coarse-scale information may
not be sufficient. Mid-range to local fine-scale information can help
characterize discriminant information of an region of interest.
However, the object size varies a lot from images to images. Different
receptive fields are needed for various object sizes and the context
around objects.  To exploit the characteristics of different scales, we
do pixel-wise classification at each hop and ensemble predicted class
probabilities from all hops. One naive approach is to use the extracted
c/w Saab features of a single hop for classification, called intra-hop
prediction. For example, c/w Saab features at hop-$i$ are of dimension
$(S_i\times S_i)\times K_i$, where $(S_i\times S_i)$ and $K_i$ represent
spatial and spectral dimensions, respectively. $N$ training images
contribute $N\times S_i\times S_i$ training samples for intra-hop
prediction. Each sample has a feature vector of dimension $K_i$. For the
$n$-th training image, all $S_i\times S_i$ pixels share the same object
label $y_n$. For the test image, the output of intra-hop prediction at
hop-$i$ is a tensor of predicted class probabilities with the same
spatial resolution of the input.  The class probability vectors are
soft decision labels of a pixel at hop-$i$, which corresponds to a
spatial region of the input image with its size equal to the receptive
field of the pixel.

Pixel-wise classification can be noisy since its training label is set
to the image label. For a dog image, only part of the image contains
discriminant dog characteristics while other part belongs to background
and has nothing to do with dogs. On one hand, the shared background of
dog and cat images will have two different labels in the training so
that its prediction of dog or cat becomes less confident. On the other
hand, the intra-hop classifier will be more confident in its predictions
in regions that are associated with a particular class uniquely. For
example, for a binary classification between a cat and a dog, contents
in background such as grass and sofa are usually common so that the
classifier is less confident in these regions as compared with the
foreground regions that contain cat/dog faces. 

\subsection{Supervised Label Smoothing (SLS)} \label{subsec:SLS}

Since the decision at each pixel is made independently of others in
pixel-level classification, there are local prediction fluctuations. We
expect consistency in predictions in a spatial neighborhood as well as
regions of a similar receptive field across hops that form a pyramid.
To reduce the prediction fluctuation, we propose a supervised label smoothing
scheme that iteratively uses predicted labels to enhance classification
performance while keeping intra-hop and inter-hop consistency. It
consists of three components: a) local graph construction; b) label
initialization; and c) iterative cross-hop label update. A similar high-level idea
was proposed in GraphHop \cite{xie2021graphhop}, yet the details
are different.

\begin{figure}[t]
\begin{center}
\includegraphics[width=0.47\linewidth]{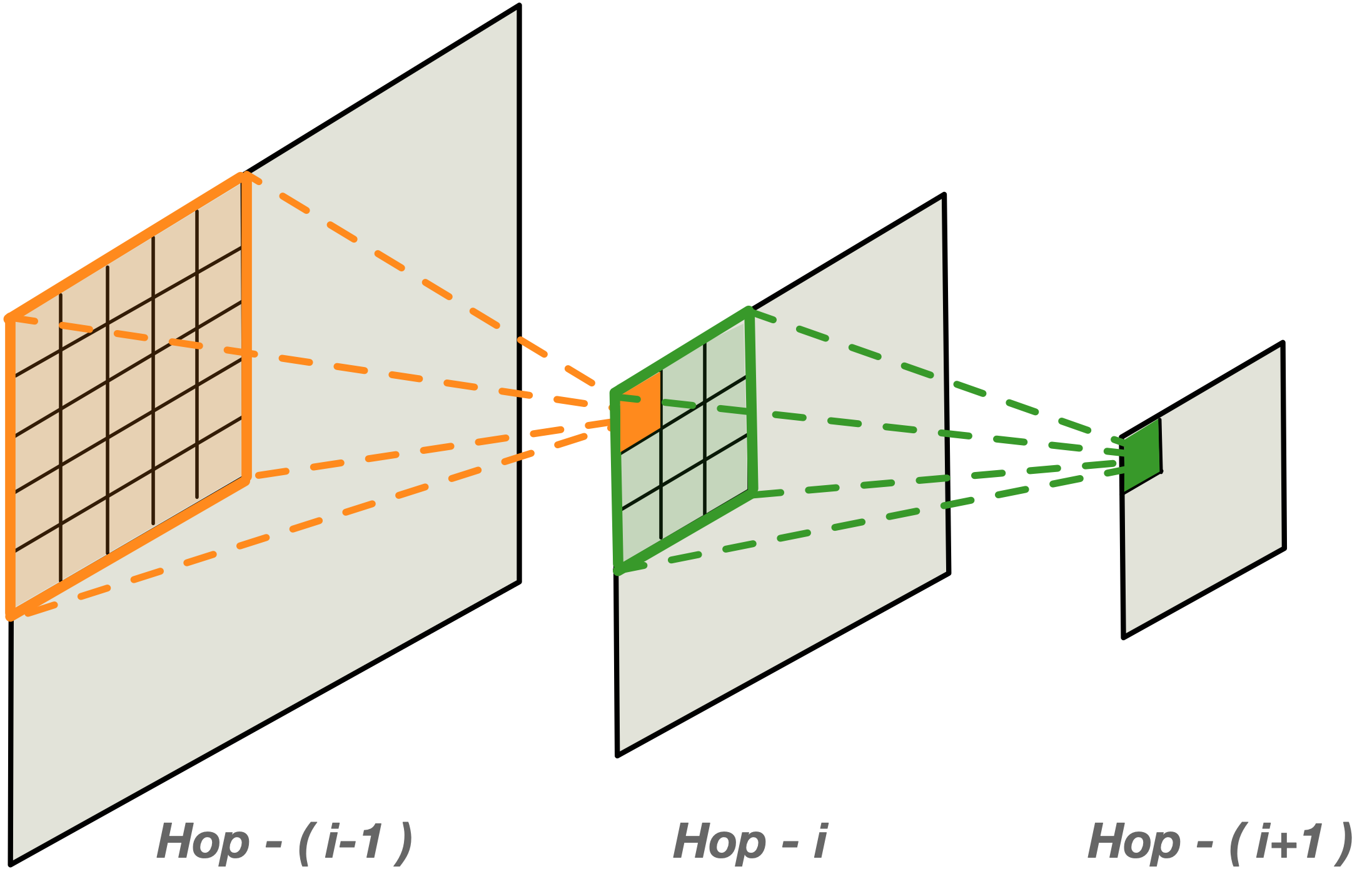}
\end{center}
\caption{Illustration of local graph construction based on c/w Saab transforms.}
\label{fig:localgraph}
\end{figure}

\paragraph{Local Graph Construction.} The tree-decomposed-structure of
multi-Hop c/w Saab transforms corresponds to a directed graph.  Each
spatial location is a node.  As shown in Fig. \ref{fig:localgraph}, we
define three node types -- child nodes, parent nodes, and sibling nodes.
Hop-$i$ features are generated by an input of size $5\times5$ through
the c/w Saab transform from hop-$(i-1)$.  The $5\times5$ locations in
hop-$(i-1)$ form $25$ child nodes for the orange node at hop-$i$. The
orange node is then covered by a window of size $3\times3$ to generate
the green node in hop-$(i+1)$. Thus, it is one of the $9$ child nodes of
the green node and the green node is the parent node of the orange one.
As to sibling nodes, they are neighboring pixels at the same hop.
Specifically, we consider eight nearest neighbors. The union of the
eight neighbors and the center pixel form a $3\times 3$ window. 

\paragraph{Label Initialization.} For the
hop-$i$ feature tensor of dimension $S_i \times S_i \times K_i$, it
consists of $S_i^2$ nodes. We treat each node at the same hop as a
sample and train a classifier to generate a soft decision as its initial
label. In the training, the image-level object class is propagated to
all nodes at all hops.  Only the c/w Saab features are used for label
prediction initially, which is given by
\begin{equation}\label{eq:LSeq1}
\boldsymbol{Z}^{(0)}_{i,u,v} = h_{i}^{(0)}(\boldsymbol{f}_{i,u,v}),
\end{equation}
where $\boldsymbol{Z}^{(0)}_{i,u,v}$ is the initial label prediction for
node $(u,v)$ at hop-$i$, $\boldsymbol{f}_{i,u,v}$ is the corresponding
c/w Saab feature vector and $h_{i}^{(0)}$ denotes the classifier.  For the hops whose child nodes have label initialization, we concatenate the c/w Saab features and the label
predicted in the previous hop to train the classifier.
That is, 
\begin{equation}\label{eq:LSeq2}
\boldsymbol{Z}^{(0)}_{i,u,v} = h_{i}^{(0)}(\boldsymbol{f}_{i,u,v} 
\oplus \boldsymbol{Z}^{(0)}_{i-1,{u},{v}}),
\end{equation}
where the initial predicted label of $M\times N$ child nodes
constructed by the local graph for node $(u,v)$ are averaged as
\begin{equation}\label{eq:LSeq3}
\boldsymbol{Z}^{(0)}_{i-1,{u},{v}} = \frac{1}{M\times N} 
\sum_{m=0}^{M-1}\sum_{n=0}^{N-1}\boldsymbol{Z}^{(0)}_{i-1,m,n}.
\end{equation}
This is different from the intra-hop prediction because of the
aggregation with child nodes and the inter-hop information is exchanged. 

%%%%%%%%%%%%%%%%%%%%%%%%%%%%%%%%%%%%%%%%%%%%%%%%%%%%%%%%%%
%%%%%%%%%%%%%%%%%%%%%%%%%%%%%%%%%%%%%%%%%%%%%%%%%%%%%%%%%%

\begin{figure}[h]
\begin{center}
\includegraphics[width=0.75\linewidth]{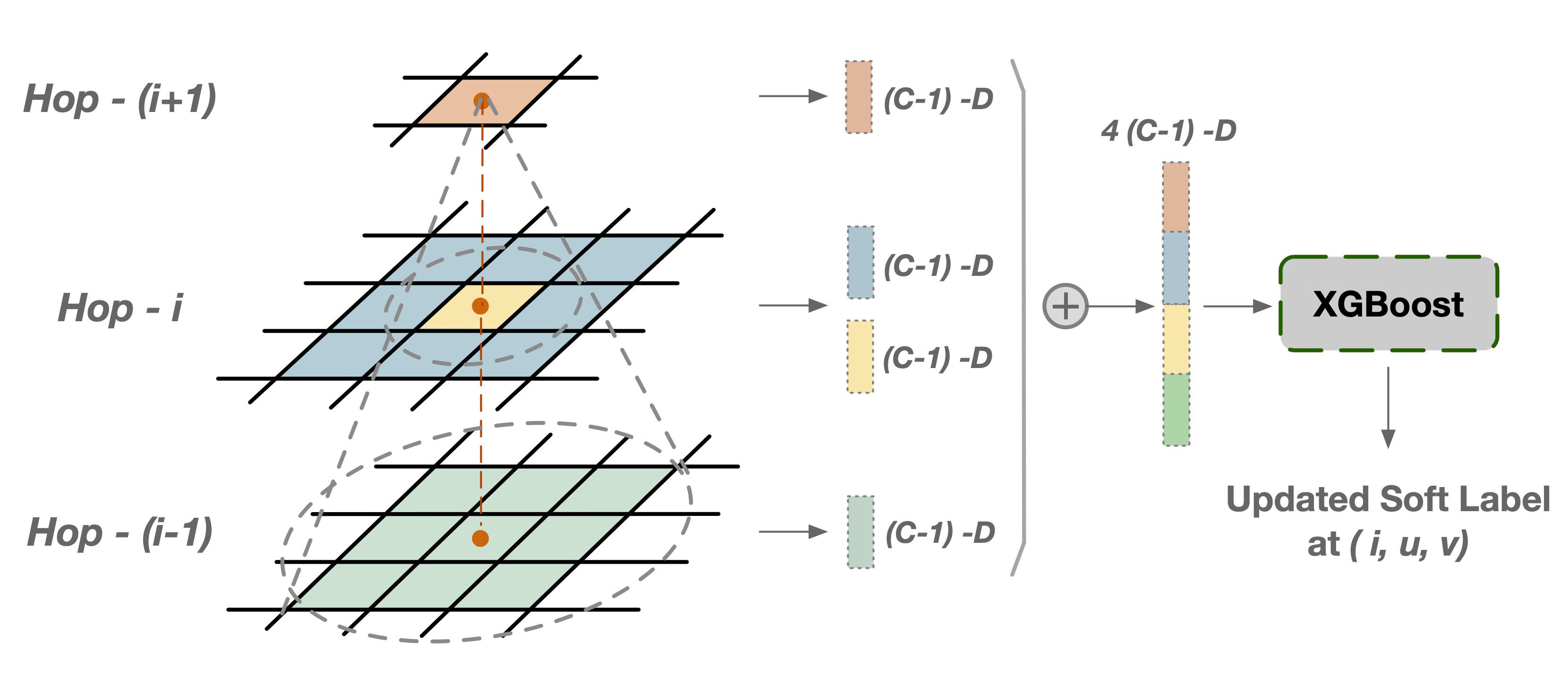}
\end{center}
\caption{Illustration of inter-hop label updates.}\label{fig:labelupdate}
\end{figure}

\paragraph{Cross-hop Label Update.} Besides the intra-hop label update
as described above, we conduct the inter-hop label update iteratively. The label is
updated from the shallow hop to the deep hop as shown in Fig.
\ref{fig:labelupdate} at each iteration.  To update the label at the
yellow location in hop-$i$, we gather the neighborhood that forms a
pyramid. The labels are averaged among sibling and child nodes in blue
and green regions, respectively. They are aggregated with labels of its
parent node and itself. For a $C$-class classification scenario, the
degree of freedom of the predicted soft label is $C-1$ so that the aggregated label
dimension is equal to $4\times(C-1)$. For example, in the Cat/Dog
confusion set with $C=2$, only the dimension corresponding to dog in the
current label is extracted. This gives the aggregated feature of
dimension 4-D. To control the model size, we do not use the c/w Saab
feature for label update. A classifier is trained against the aggregated
label dimension. The output soft decisions are the updated labels at
iteration $k$ in form of
\begin{equation}\label{eq:LSeq4}
\boldsymbol{Z}^{(k)}_{i,u,v} = h_{i}^{(k-1)}(\boldsymbol{Z_{agg}}^{(k-1)}_{i,u,v}),
\end{equation}
where $\boldsymbol{Z_{agg}}^{(k-1)}_{i,u,v}$ denotes the aggregated features
of the cross-hop neighborhood. 

%%%%%%%%%%%%%%%%%%%%%%%%%%%%%%%%%%%%%%%%%%%%%%%%%%%%%%%%%%
%%%%%%%%%%%%%%%%%%%%%%%%%%%%%%%%%%%%%%%%%%%%%%%%%%%%%%%%%%
\section{Experiments}\label{sec:experiments}

\subsection{Experimental Setup}\label{subsec:settings}

We evaluate the classification performance of E-PixelHop using the
CIFAR-10 dataset that contains 10 classes of tiny color images of
spatial resolution $32\times 32$. There are 50,000 training images and
10,000 test images. The hyper parameters of the PixelHop++ feature
extraction architecture are listed in Table \ref{tab:setting}. The
spectral dimensions are decided by energy thresholds in the c/w Saab
transform. 

For baseline classification in stage-1 and confusion class resolution in
stage-2, we adopt different pixel-based prediction schemes. For stage-1, pixel-based prediction is conducted among three output feature maps, namely, hop-2 after max-pooling, hop-3 and hop-4. The SLS method without cross-hop label update is performed among these three hops. For stage-2, besides hop-3 and hop-4, we consider hop-2 without max-pooling for higher resolution to leverage
more local information in distinguishing confusion classes. The full SLS method is conducted as the pixel-based prediction in stage-2. Yet, hop-1
features are not used since their receptive field is too small. We use
the XGBoost (extreme gradient boosting) classifier for all
classification tasks in E-PixelHop. 

By following \cite{simonyan2014very}, we adopt data augmentation in both
training and test for performance improvement. It includes original
input images, random squared/rectangular croppings, contrast
manipulation, horizontal flipping. Eight variants of each input are
created. The Lanczos interpolation is applied to cropped images to
resize them back to the $32\times 32$ resolution. In the test phase,
classification soft decisions are averaged among the eight variants to
get the final score for a test image. 

%%%%%%%%%%%%%%%%%%%%%%%%%%%%%%%%%%%%%%%%%%%%%%%%%%%%%%%%%%
\begin{table}[!htbp]
\begin{center}
\caption{Hyper parameters of the PixelHop++ feature extraction architecture.}
\label{tab:setting}
\begin{tabular}{l|l|l|c|c}
\hline
\multirow{2}{*}{} & \multicolumn{2}{c|}{Filter}     & \multicolumn{2}{c}{Output Feature} \\ \cline{2-5} 
                  & Window Size        & Stride     & Spatial            & Spectral (P/Q) \\ \hline
Hop-1             & $5\times5$ (pad 2) & $1\times1$ & $32\times32$       & 24 / 22             \\
Max-pooling 1     & $3\times3$         & $2\times2$ & $15\times15$       & 24 / 22             \\
Hop-2             & $5\times5$ (pad 2) & $1\times1$ & $15\times15$       & 144 / 114            \\
Max-pooling 2     & $3\times3$         & $2\times2$ & $7\times7$         & 144 / 114            \\
Hop-3             & $3\times3$         & $1\times1$ & $5\times5$         & 203 / 174            \\
Hop-4             & $3\times3$         & $1\times1$ & $3\times3$         & 211 / 185            \\ \hline
\end{tabular}
\end{center}
\end{table}
%%%%%%%%%%%%%%%%%%%%%%%%%%%%%%%%%%%%%%%%%%%%%%%%%%%%%%%%%%

\subsection{Effects of Color Decomposition}\label{subsec:effect_pqr}

E-PixelHop conducts classification in P and Q channels separately and
then ensembles the classification results as shown in Fig.
\ref{fig:baseline}. We compare different ways to handle color channels
for the baseline classifier in Table \ref{tab:comparePQ}. Since the Q
channel contains much less energy than the P channel, its classification
performance is worse. The ensemble of P and Q channels outperforms the P
channel alone by 3.46\% and 2.83\% in top-1 and top-2 accuracies,
respectively. It shows that P and Q channels are complementary to each
other. 

%%%%%%%%%%%%%%%%%%%%%%%%%%%%%%%%%%%%%%%%%%%%%%%%%%%%%%%%%%
\begin{table}[t]
\centering
\caption{Comparison of test accuracy (\%) in E-PixelHop baseline using 
different color channels}\label{tab:comparePQ}
\begin{tabular}{cc|cc}
\hline
P channel   & Q channel     & Top-1             & Top-2 \\ \hline
\checkmark  &               & 70.26             & 84.92 \\
            & \checkmark    & 58.8              & 76.13 \\
\checkmark  & \checkmark    & \textbf{73.72}    & \textbf{87.75} \\ \hline
\end{tabular}
\end{table}
%%%%%%%%%%%%%%%%%%%%%%%%%%%%%%%%%%%%%%%%%%%%%%%%%%%%%%%%%%

\subsection{Confusion Sets Ranking}\label{subsec:effect_confuse}

\begin{figure}[t]
\begin{center}
\includegraphics[width=0.45\linewidth]{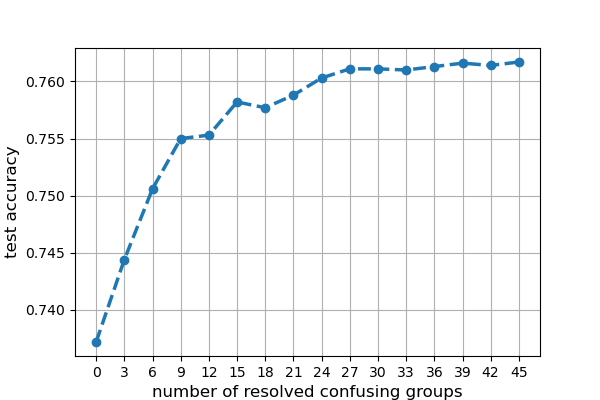}
\end{center}
\caption{The plot of CIFAR-10 test accuracy (\%) as a function of 
the cumulative number of resolved confusion sets.}
\label{fig:plot1}
\end{figure}

For a 10-class classification problem, we may have at most 45 confusion
sets for one-versus-one competition.  Two images form a confusion pair
if they are the top-2 candidates in the baseline decision.  The image
number in each confusion set varies. For the second stage
classification, we give a higher priority to confusion sets that have a
large number of pairs and plot the test accuracy as a function of the
total resolved confusing sets in Fig.  \ref{fig:plot1}.  The test
accuracy increases as more confusion sets are resolved and saturates
after 27 sets are taken care of. 

%%%%%%%%%%%%%%%%%%%%%%%%%%%%%%%%%%%%%%%%%%%%%%%%%%%%%%%%%%
%%%%%%%%%%%%%%%%%%%%%%%%%%%%%%%%%%%%%%%%%%%%%%%%%%%%%%%%%%

\subsection{Effects of Supervised Label Smoothing (SLS)}\label{subsec:effect_LS}

To demonstrate the power of SLS, we study the binary classification
problem and show the results in Table \ref{tab:tab2} for four frequent
confusion sets: Cat vs Dog, Airplane vs Ship, Automobiles vs Truck, and
Deer vs Horse.  The training set includes 5,000 images while the test
set contains 1,000 images from each of the two classes.  For pixel-based
classification, we examine test accuracies of intra-hop prediction and
the addition of supervised label smoothing (SLS), where results of P and
Q channels are ensembled. We see from the table that the addition of SLS
outperforms that without SLS by 2\%$\sim$3\% in all four confusing sets. 

%%%%%%%%%%%%%%%%%%%%%%%%%%%%%%%%%%%%%%%%%%%%%%%%%%%%%%%%%%
\begin{table}[!htbp]
\centering
\caption{Comparison of image-level test accuracy (\%) for four
confusion sets with and without SLS.}
\label{tab:tab2}
\begin{tabular}{lcc}
\hline
Confusing Group     & intra-hop only & intra-hop and SLS \\ \hline
Cat vs Dog          & 76.45         & 79.1         \\
Airplane vs Ship    & 92.1          & 93.75         \\
Automobile vs Truck & 89.8          & 92.95         \\
Deer vs Horse       & 87.8          & 90.95         \\ \hline
\end{tabular}
\end{table}
%%%%%%%%%%%%%%%%%%%%%%%%%%%%%%%%%%%%%%%%%%%%%%%%%%%%%%%%%%
%%%%%%%%%%%%%%%%%%%%%%%%%%%%%%%%%%%%%%%%%%%%%%%%%%%%%%%%%%
\begin{figure*}[t]
\begin{center}
\includegraphics[width=0.97\linewidth]{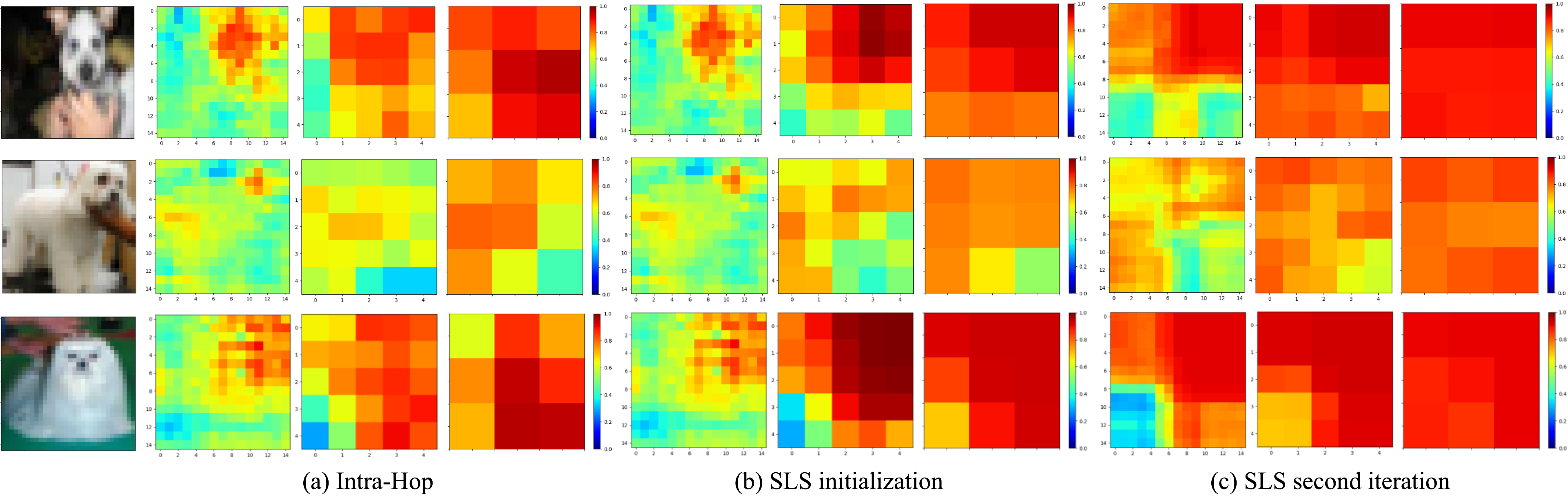}
\end{center}
\caption{Illustration of intermediate label maps for different methods,
where each subfigure shows label heatmaps in hop-2, hop-3 and hop-4
(from left to right).} \label{fig:LS_heatmap}
\end{figure*}
%%%%%%%%%%%%%%%%%%%%%%%%%%%%%%%%%%%%%%%%%%%%%%%%%%%%%%%%%%

To understand the behavior of SLS furthermore, we show intermediate label
maps in heat maps from hop-2 to hop-4 for three Dog images in Fig.
\ref{fig:LS_heatmap} with three schemes: a) intra-hop prediction, b) one iteration of SLS label initialization only and c) SLS with another iteration of cross-hop label update. The heatmaps indicate the
confidence score at each pixel location for the Dog class. For the
hop-2 heatmaps, regions composed by the dog face have higher Dog
confidence than other regions. By comparing Figs.
\ref{fig:LS_heatmap}(a) and (b), we see an increase in the highest
confidence level in hop-3 using one round of SLS label initialization. Clearly, SLS makes
label maps more distinguishable. Furthermore, heatmaps are more
consistent across hops in Fig. \ref{fig:LS_heatmap}(b). After one more
SLS iteration of cross-hop update, the confidence scores become more homogeneous in a
neighborhood as shown in Fig. \ref{fig:LS_heatmap}(c), where more
locations agree with each other and with the ground truth label. 
%%%%%%%%%%%%%%%%%%%%%%%%%%%%%%%%%%%%%%%%%%%%%%%%%%%%%%%%%%
\begin{figure*}[t]
\begin{center}
\includegraphics[width=1.0\linewidth]{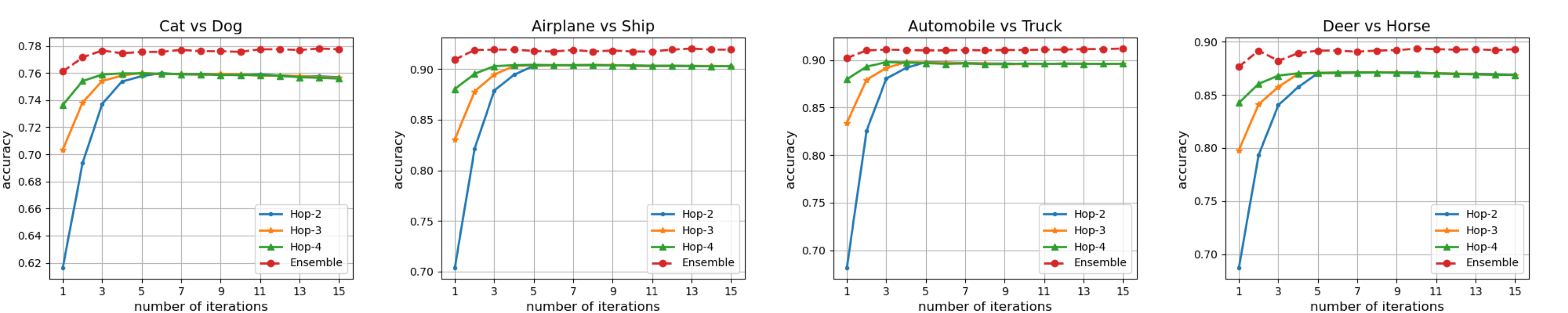}
\end{center}
\caption{Pixel-level test accuracy in hop-2, hop-3 and hop-4 and the
image-level ensemble as a function of iteration numbers for 
P-channel images only.} \label{fig:LS_curve}
\end{figure*}
%%%%%%%%%%%%%%%%%%%%%%%%%%%%%%%%%%%%%%%%%%%%%%%%%%%%%%%%%%

Since SLS is an iterative process, we show the pixel-level accuracy and
the final image-level accuracy as a function of the iteration number for
the four confusion sets discussed earlier in Fig. \ref{fig:LS_curve}.
The first point is the initial label in SLS. We see that the pixel-level
accuracy at each hop increases rapidly in first several iterations.
This is especially obvious in shallow hops. Then, these curves converge
at a certain level. In the experiments, we set the maximum number of
iterations to \textit{num\_iter} = 3 for the one-versus-one confusion
set resolution to avoid over-smoothing.  In the 10-class baseline, we
set \textit{num\_iter} = 1 (i.e. only the initialization of SLS) by
taking both inter-hop guidance and model complexity into account.

Although the pixel-level accuracy saturates at the same level for all
three hops, the ensemble performance (in red dashed line) is better than
that of each individual hop as shown in Fig.  \ref{fig:LS_curve}. As the
iteration number increases, the ensemble result also increases and
saturates in 3 to 5 iterations although the increment is smaller than
that of the pixel-level accuracy. 
%%%%%%%%%%%%%%%%%%%%%%%%%%%%%%%%%%%%%%%%%%%%%%%%%%%%%%%%%%
\begin{table*}[b]
\centering
\caption{Ablation study of E-PixelHop's components for CIFAR-10}
\label{tab:ablation}
\begin{tabular}{c|cc|cc|cc|c}
\hline
\multirow{2}{*}{Augmentation} & \multicolumn{2}{c|}{Color Channels} & \multicolumn{2}{c|}{Baseline} & \multicolumn{2}{c|}{2-nd pipeline} & \multirow{2}{*}{Test Accuracy} \\ \cline{2-7}
                              & P channel        & Q channel       & intra-hop     & SLS\_init    & intra-Hop & SLS       & \\ \hline
                              & \checkmark        & \checkmark       &               & \checkmark    &         &           & 68.69 \\
\checkmark                     & \checkmark        &                 &               & \checkmark    &         &           & 70.26 \\
\checkmark                     & \checkmark        & \checkmark       &               & \checkmark    &         &           & \textbf{73.72} \\
\checkmark                     & \checkmark        & \checkmark       & \checkmark     &              &           &           & 72.29  \\
\checkmark                     & \checkmark        & \checkmark       &               & \checkmark    & \checkmark &           & 72.79 \\
\checkmark                     & \checkmark        & \checkmark       &               & \checkmark    &           & \checkmark & \textbf{76.18} \\ 
\checkmark                     & \checkmark        & \checkmark       & \checkmark     &              &           & \checkmark & 75.54 \\ \hline

\end{tabular}%
%}
\end{table*}
%%%%%%%%%%%%%%%%%%%%%%%%%%%%%%%%%%%%%%%%%%%%%%%%%%%%%%%%%%

\subsection{Ablation Study and Performance Benchmarking}\label{subsec:benchmark}

The ablation study of adopting various components in E-PixelHop is
summarized in Table \ref{tab:ablation}, where the image-level test
accuracy for CIFAR-10 is given in the last column. The study includes:
data augmentation, color channel decomposition, the methods of
pixel-level classification in the baseline and in confusion set
resolution, respectively. The first four rows show the baseline
performance without confusion set resolution while the last three rows
have both.  For pixel-level classification, the intra-hop column does
not have label smoothing while the SLS\_init column has SLS
initialization only. Moreover, the pixel-level classification in the
baseline does not use augmented images to train classifiers by
considering the time and model complexity.  Data augmentation is only
used for the meta classification in the baseline and one-versus-one competition in confusion set resolution. 
The E-PixelHop baseline achieves test accuracy of 73.72\% using both P/Q
channels with SLS initialization for pixel-level classification and data
augmentation when training the meta classifier. By further adding the
$2^{nd}$ pipeline with three SLS iterations in pixel-level
classification, E-PixelHop achieves a test accuracy of 76.18\% as shown
in the sixth row, where all 45 confusing sets are processed. 

We compare the performance of six methods in Table \ref{tab:benchmark}.
They are modeified LeNet-5~\cite{lecun1998gradient}, PixelHop
\cite{chen2020pixelhop}, PixelHop$^+$ \cite{chen2020pixelhop} and
PixelHop++ \cite{chen2020pixelhop++}, the E-PixelHop baseline, and the
complete E-PixelHop. To handle color images, we modify the LeNet-5
network architectures slightly, whose hyper parameters are given in
Table \ref{tab:lenet5}. Both the E-PixelHop baseline and the complete
E-PixelHop outperform other benchmarking methods.  An improvement of
9.37\% over PixelHop++ and an improvement of 3.52\% over PixelHop$^+$
are observed, respectively. 

%%%%%%%%%%%%%%%%%%%%%%%%%%%%%%%%%%%%%%%%%%%%%%%%%%%%%%%%%%
\begin{table}[h]
\centering
\caption{Comparison of testing accuracy (\%) of LeNet-5, PixelHop, 
PixelHop+ and PixelHop++ for CIFAR-10.}\label{tab:benchmark}
\begin{tabular}{lc}
\hline
                                        & Test Accuracy (\%) \\ \hline
Lenet-5                                 & 68.72              \\ \hline
PixelHop \cite{chen2020pixelhop}        & 71.37              \\
PixelHop$^+$ \cite{chen2020pixelhop}    & 72.66              \\
PixelHop++ \cite{chen2020pixelhop++}    & 66.81              \\ \hline
E-PixelHop Baseline (Ours)              & 73.72              \\
E-PixelHop  (Ours)                      & \textbf{76.18}     \\ \hline
\end{tabular}
\end{table}
%%%%%%%%%%%%%%%%%%%%%%%%%%%%%%%%%%%%%%%%%%%%%%%%%%%%%%%%%%

%%%%%%%%%%%%%%%%%%%%%%%%%%%%%%%%%%%%%%%%%%%%%%%%%%%%%%%%%%
\begin{table}[h]
\centering
\caption{Hyper parameters of the modified LeNet-5 network as 
compared with those of the original LeNet-5.}\label{tab:lenet5}
\begin{tabular}{lcc}
\hline
                   & Original LeNet-5 & Modified LeNet-5 \\ \hline
Conv-1 Kernel Size & 5x5x1            & 5x5x3            \\
Conv-1 Kernel No.  & 6                & 32               \\
Conv-2 Kernel Size & 5x5x6            & 5x5x32           \\
Conv-2 Kernel No.  & 16               & 64               \\
FC-1               & 120              & 200              \\
FC-2               & 84               & 100              \\
Output Layer       & 10               & 10               \\ \hline
\end{tabular}
\end{table}
%%%%%%%%%%%%%%%%%%%%%%%%%%%%%%%%%%%%%%%%%%%%%%%%%%%%%%%%%%

\section{Conclusion and Future Work}\label{sec:conclude}

An enhanced SSL-based object classification method called E-PixelHop was
proposed in this work. It has two classification stages: multi-class
classification in the first stage and the confusion set resolution in
the second stage. Supervised label smoothing (SLS) was proposed to
enhance the performance and ensure consistency of intra-hop and
inter-hop pixel-level predictions. Effectiveness of SLS shows the
importance of prediction agreement at different scales. Furthermore, it
is important to handle confusing classes carefully so that the overall
classification performance can be boosted. 

For the second-stage classification, all training images in the
confusion set are used to train the one-versus-one model in the current
E-PixelHop. In the future, it is worthwhile to investigate boosting with
hard cases mining and learn from errors.  Furthermore, not every pixel
in an image is discriminant and salient. We should find a way to focus
on discriminant regions by introducing an attention mechanism. It is
also desired to generalize our work to the recognition of a larger
number of classes (e.g., CIFAR-100) and higher image resolution (e.g.,
the ImageNet).

%\clearpage
\bibliographystyle{ieeetr}
\bibliography{ref.bib}

\end{document}